\pdfoutput=1

\documentclass[11pt]{article}

\usepackage[]{ACL}

\usepackage{times}
\usepackage{latexsym}

\usepackage[T1]{fontenc}

\usepackage[utf8]{inputenc}

\usepackage{microtype}

\usepackage{inconsolata}

\usepackage{graphicx} 
\usepackage{algorithm}
\usepackage{algorithmic}
\usepackage{amsmath}
\usepackage{amsthm}
\usepackage{booktabs}
\usepackage{color}
\usepackage{multirow}
\usepackage{array}
\usepackage{bbm}
\usepackage{arydshln}
\usepackage{dsfont}
\usepackage[utf8]{inputenc}
\usepackage{cleveref}
\crefname{section}{§}{§§}
\Crefname{section}{§}{§§}
\usepackage{makecell}
\usepackage{bbding}
\usepackage{amssymb}
\usepackage{pifont}%

\newcommand*{\img}[1]{%
    \raisebox{-.2\baselineskip}{%
        \includegraphics[
        scale=0.55,
        keepaspectratio,
        ]{#1}%
    }%
}

\title{PromptNER: Prompt Locating and Typing for Named Entity Recognition}

\author{
Yongliang Shen$^{1}$, Zeqi Tan$^{1}$, Shuhui Wu$^{1}$, Wenqi Zhang$^{1}$,\\ 
\textbf{Rongsheng Zhang$^{2}$, Yadong Xi$^{2}$, Weiming Lu$^{1\dagger}$, Yueting Zhuang$^{1}$}\\
$^{1}$College of Computer Science and Technology, Zhejiang University \\
$^{2}$Fuxi AI Lab, NetEase Inc.\\
\texttt{syl@zju.edu.cn}
}

\begin{document}

\maketitle
\renewcommand{\thefootnote}{\fnsymbol{footnote}}
\footnotetext[2]{\;Corresponding author.}
\renewcommand{\thefootnote}{\arabic{footnote}}

\begin{abstract}

Prompt learning is a new paradigm for utilizing pre-trained language models and has achieved great success in many tasks. To adopt prompt learning in the NER task, two kinds of methods have been explored from a pair of symmetric perspectives, populating the template by enumerating spans to predict their entity types or constructing type-specific prompts to locate entities. However, these methods not only require a multi-round prompting manner with a high time overhead and computational cost, but also require elaborate prompt templates, that are difficult to apply in practical scenarios. In this paper, we unify entity locating and entity typing into prompt learning, and design a dual-slot multi-prompt template with the position slot and type slot to prompt locating and typing respectively. Multiple prompts can be input to the model simultaneously, and then the model extracts all entities by parallel predictions on the slots. To assign labels for the slots during training, we design a dynamic template filling mechanism that uses the extended bipartite graph matching between prompts and the ground-truth entities. We conduct experiments in various settings, including resource-rich flat and nested NER datasets and low-resource in-domain and cross-domain datasets. Experimental results show that the proposed model achieves a significant performance improvement, especially in the cross-domain few-shot setting, which outperforms the state-of-the-art model by +7.7\% on average\footnote{\;Our code will be available at \url{https://github.com/tricktreat/PromptNER}.}.

\end{abstract}

\section{Introduction}

Named entity recognition (NER) is a fundamental task in natural language processing that aims to identify specific types of entities in free text, such as person, location, and organization.
Traditional sequence labeling methods \citep{ma-hovy-2016-end} have difficulty coping with nested entities, and recent works have transformed NER into other paradigms such as reading comprehension \citep{li-etal-2020-unified, shen-etal-2022-parallel}, set prediction \citep{ijcai2021-542, ijcai2022-0613} and sequence generation \citep{tanl,yan-etal-2021-unified-generative,lu-etal-2022-unified}. However, low-resource and cross-domain problems in practical scenarios still pose a great challenge to NER models.

\begin{figure}[t!]
    \centering
    \includegraphics[width=\linewidth]{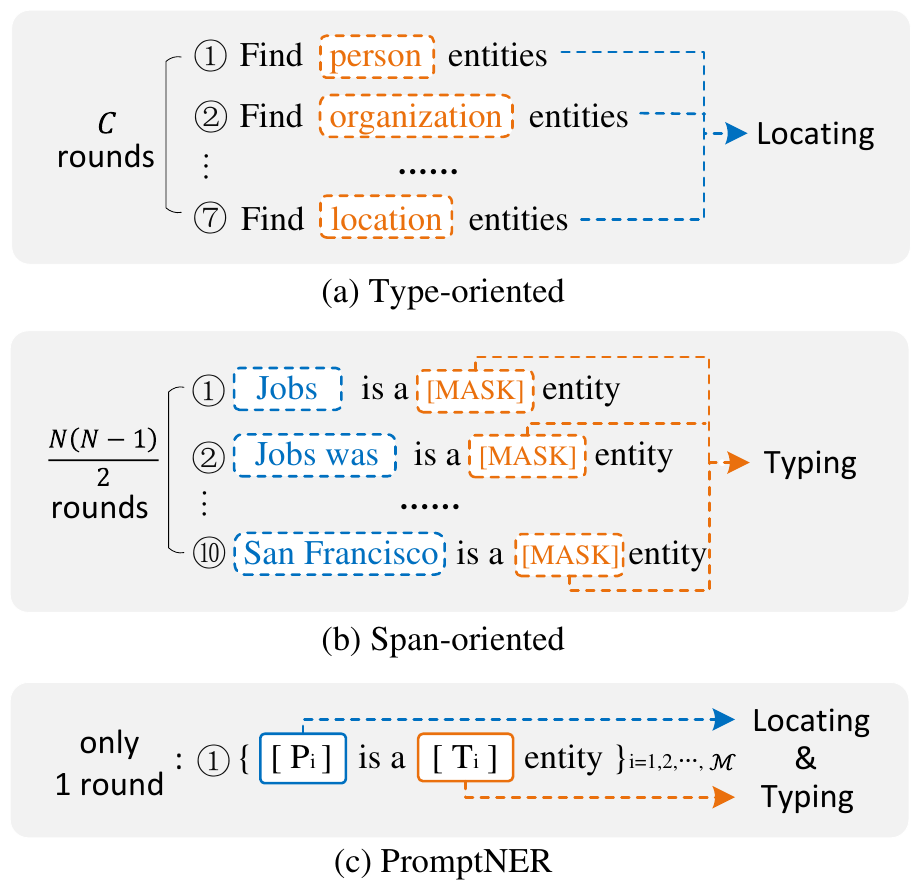}
    \caption{A comparison of the type-oriented (a) and span-oriented (b) prompt learning with the proposed PromptNER (c). $C$, $N$ and $\mathcal{M}$ denote the number of entity types, words and prompts, respectively.}
    \label{fig:comparison}
\end{figure}

Recently prompt learning \citep{https://doi.org/10.48550/arxiv.2107.13586, GPTUnderstand, li-liang-2021-prefix, lester-etal-2021-power} has received a lot of interest because of its excellent performance and data efficiency, and has been adopted in many classification and generation tasks \citep{gao-etal-2021-making, schick-schutze-2021-just, ding2021prompt, wu2022towards}.
Prompt learning converts downstream tasks into language modeling tasks, where cloze questions are constructed as prompts to guide pre-trained language models to fill in the blanks. 
However, named entity recognition is a token-level tagging task, and it is difficult to apply prompt-based learning on NER directly \citep{https://doi.org/10.48550/arxiv.2107.13586}.
\citet{cui-etal-2021-template} proposes the template-based method, which constructs prompts for each potential entity span and then separately predicts their entity types.
For example, given an input \textit{``Jobs was born in San Francisco''}, \citet{cui-etal-2021-template} enumerates each span to populate \texttt{[X]} of the template \textit{``}\texttt{[X]}\textit{ is a }\texttt{[MASK]}\textit{ entity''}, and then determines the type of the filled span based on the prediction on the \texttt{[MASK]} slot.
In contrast to entity typing over the enumerated spans, some methods \citep{li-etal-2020-unified, liu2022qaner} design prompt templates from a symmetric perspective. They construct prompts for each entity type and then guide the model to locate specific types of entities.
For example, \citet{liu2022qaner} constructs the prompt \textit{``What is the location?''} for the \texttt{LOC} type, and then predicts all \texttt{LOC} entities in the sentence, e.g., \textit{``San Francisco''}.
We group these two types of methods into span-oriented and type-oriented prompt learning. As shown in Figure \ref{fig:comparison}, they construct prompts based on the entity span or entity type, and then perform entity typing or entity locating. However, both groups of methods require multiple rounds of prompting. For an input with $N$ words and $C$ pre-fixed types, type-oriented and span-oriented prompt learning require $C$ and ${N(N-1)}/{2}$ predictions, respectively. Moreover, each round of prediction is independent of the other, ignoring the latent relationships between different entities.

Different from the above methods that either perform multiple rounds of entity typing or entity locating through prompting, in this paper, we propose a prompt learning method for NER (\textbf{PromptNER}) that unifies entity locating and entity typing into one-round prompt learning. Specifically, we design the position slot \texttt{[P]} and the type slot \texttt{[T]} in the prompt template, which are used for prompting entity locating and typing accordingly. This manner is enumeration-free for entity span or entity type, and can locate and classify all entities in parallel, which improves the inference efficiency of the model.
Since the correspondence between prompts and entities cannot be specified in advance, we need to assign labels to the slots in the prompts during training. 
Inspired by \citet{10.1007/978-3-030-58452-8_13}, we treat the label assignment process as a linear assignment problem and perform bipartite graph match problem between the prompts and the entities. We further extend the traditional bipartite graph matching and design a one-to-many dynamic template filling mechanism so that an entity can be predicted by multiple prompts, which can improve the utilization of prompts. To summarize, our main contributions are as follows:

\begin{itemize}
\item We unify entity locating and entity typing for NER in prompt learning by filling both position and type slots in the dual-slot multi-prompt template. Our model eliminates the need to enumerate entity types or entity spans and can predict all entities in one round.
\item For the model training, we design a dynamic template filling mechanism to assign labels for the position and type slots by an extended bipartite graph matching.
\item We conduct experiments in a variety of settings, and we achieve significant performance improvements on both standard flat and nested NER datasets. In the cross-domain few-shot setting, our model outperforms the previous state-of-the-art models by +7.7\% on average. 
\end{itemize}

\section{Related Work}

\subsection{Named Entity Recognition}
Named Entity Recognition (NER) is a basic task of information extraction \citep{tjong-kim-sang-de-meulder-2003-introduction, wadden-etal-2019-entity, 10.1145/3442381.3449895, tan-etal-2022-query}.
Current named entity recognition methods can be divided into four categories, including tagging-based, span-based, hypergraph-based, and generative-based methods. Traditional tagging-based methods \citep{ma-hovy-2016-end} predict a label for each word, which is difficult to cope with nested entities. Some works propose various strategies for improvement. For example, \citet{alex-etal-2007-recognising} and \citet{ju-etal-2018-neural} use cascading or stacked tagging layers, and \citet{wang-etal-2020-pyramid} designs the tagging scheme with a pyramid structure. The span-based methods \citep{sohrab-miwa-2018-deep} model NER as a classification task for spans directly, with the inherent ability to recognize nested entities. Due to the high cost of exhausting all spans, \citet{zheng-etal-2019-boundary} and \citet{shen-etal-2021-locate} propose boundary-aware and boundary-regression strategies based on span classification, respectively. Some other methods \citep{ yu-etal-2020-named, li2022unified} perform classification on inter-word dependencies or interactions, which are essentially span classification, and can also be considered as span-based methods.
The generative-based methods \citep{yan-etal-2021-unified-generative, lu-etal-2022-unified, zhang-etal-2022-de} are more general. They model the NER task as a sequence generation task that can unify the prediction of flat and nested entities.

In addition, some works focus on the NER task in practical settings, including the few-shot NER \citep{ding-etal-2021-nerd} and the cross-domain NER \citep{Liu_Xu_Yu_Dai_Ji_Cahyawijaya_Madotto_Fung_2021}. For example, \citet{chen-etal-2021-data} and \citet{zhou-etal-2022-melm} design data augmentation methods augment labeled data on low-resource domains. Some works \citep{DBLP:journals/corr/abs-2008-10570, wiseman-stratos-2019-label} use the instance learning to perform a nearest neighbor search based on entity instances or token instances, and others \citep{ding-etal-2021-nerd, huang-etal-2021-shot} use prototype networks at the token level or span level to handle such low-resource settings. 

\subsection{Prompt Learning}

Prompt learning constructs prompts by injecting the input into a designed template, and converts the downstream task into a fill-in-the-blank task, then allows the language model to predict the slots in the prompts and eventually deduce the final output. Due to the data efficiency, prompt learning is currently widely used for many classification and generation tasks \citep{shin-etal-2020-autoprompt, gao-etal-2021-making, schick-schutze-2021-just, schick-schutze-2021-exploiting, ding2021prompt}. Some works investigate prompt learning on the extraction tasks. \citet{cui-etal-2021-template} first applies prompt learning to NER. It proposes a straightforward way to construct separate prompts in the form of \textit{``}\texttt{[X]}\textit{ is a }\texttt{[MASK]}\textit{ entity''} by enumerating all spans. The model then classifies the entities by filling the \texttt{[MASK]} slot.
Since these methods need to construct templates and perform multiple rounds of inference, \citet{ma-etal-2022-template} proposes a template-free prompt learning method using the mutual prediction of words with the same entity type. However, it requires constructing sets of words of the same entity type, which is difficult in low-resource scenarios. 
\citet{lee-etal-2022-good} introduces demonstration-based learning in low-resource scenarios, they concatenate demonstrations in the prompts, including entity-oriented demonstrations and instance-oriented demonstrations.
Another class of query-based methods \citep{li-etal-2020-unified, mengge-etal-2020-coarse, liu2022qaner} can also be categorized as prompt learning. 
In contrast to the above methods, they construct a type-related prompt (query), e.g. ``\textit{Who is the person ?}", and then lets the model locate all \texttt{PER} entities in the input. 
Different from all of the above, we unify entity locating and entity typing in prompt learning, and predict all entities in one round using a dual-slot multi-prompt template.

\section{Method}

\begin{figure*}
    \centering
    \includegraphics[width=\linewidth]{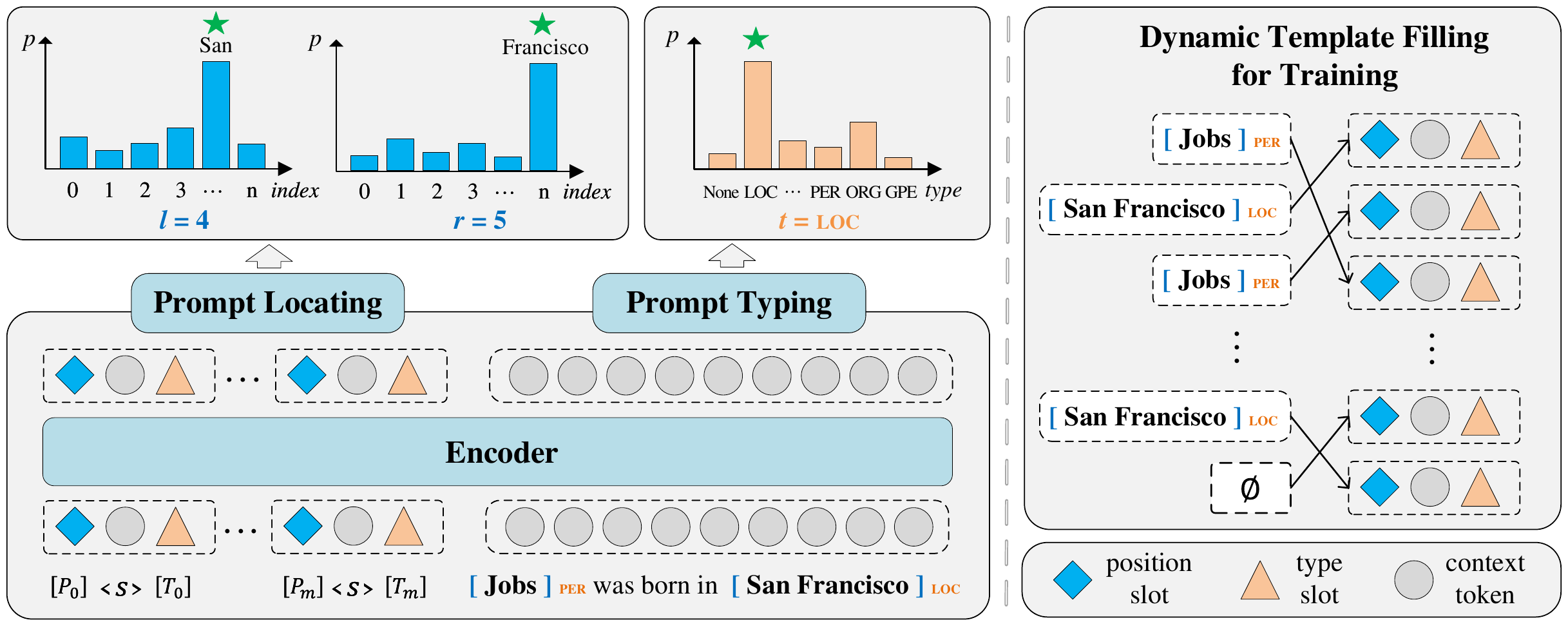}
    \caption{An overview of PromptNER. The left part describes the model's inference process and the right part describes the dynamic template filling mechanism during training. The model takes a dual-slot multi-prompt sequence as input and fills in the position slot `\img{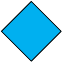}' and type slot `\img{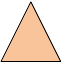}'  by prompt locating and prompt typing.}
    \label{fig:model}
\end{figure*}

In this section, we first introduce the task formulation in \Cref{3.1}, and then describe our method. The overview of the PromptNER is shown in Figure \ref{fig:model}, and we will introduce the prompt construction in \Cref{3.2} and the model inference in \Cref{3.3}, including the encoder and the entity decoding module. The training of the model requires assigning labels to the slots of the prompt, and we will introduce the dynamic template filling mechanism in \Cref{3.4}.

\subsection{Task Formulation}
\label{3.1}

Following \citet{cui-etal-2021-template} and \citet{lee-etal-2022-good}, we transform the NER task into a fill-in-the-blank task. Given a sentence $X$ of length $N$, we fill a fixed number $\mathcal{M}$ of prompts and $X$ into a predefined template to construct the complete input sequence $\mathcal{T}$. The model then fills the position slots \texttt{[P]} and type slots \texttt{[T]} of all prompts and decodes the named entities in the sentence.

\subsection{Prompt Construction}
\label{3.2}

Different from the previous methods, we unify entity locating and entity typing into one-round prompt learning. Therefore, we have two improvements in prompt construction. First, each prompt has two slots, entity position slot and entity type slot, which are used for entity locating and entity typing respectively. Second, our model fills slots for a predefined number of prompts simultaneously and extracts all entities in a parallel manner.
Specifically, the constructed input sequence consists of two parts: $\mathcal{M}$ prompts and the input sentence $X$. By default, each prompt has only two tokens: a position slot \texttt{[P]} and a type slot \texttt{[T]}. For a sentence $X=$\textit{``Jobs was born in San Francisco''}, the default dual-slot multi-prompt input sequence can be represented as:
\begin{equation*}
\mathcal{T} =
\begin{aligned}
&\left\{\left[ \texttt{P}_{i}\right] \textit { is a }\left[\texttt{T}_{i}\right] \textit { entity }\right\}_{i=1,2, \cdots, \mathcal{M}} \\
& [\texttt{CLS}] \textit { Jobs was born in San Francisco.}
\end{aligned} \\
\end{equation*}

\noindent where `` $\left[\texttt{P}_{i}\right] \textit{ is a }\left[\texttt{T}_{i}\right] \textit{ entity }$'' is 
 the $i$-th prompt, \texttt{[P$_{i}$]} and \texttt{[T$_{i}$]} denote its position and type slots and $\mathcal{M}$ denotes the number of prompts. Following \citet{lester-etal-2021-power, gao-etal-2021-making}, we also experiment with soft templates by replacing concrete contextual words with learnable tokens. In \Cref{template} we compare the performance of the model using different templates.

\subsection{Prompt Locating and Typing}
\label{3.3}

With the input sequence $\mathcal{T}$ filled with the sentence $X$ and $\mathcal{M}$ prompts, the model decodes the entities by filling the position slots \texttt{[P$_{i}$]$_{i=1,2, \cdots, \mathcal{M}}$} and type slots \texttt{[T$_{i}$]$_{i=1,2, \cdots, \mathcal{M}}$} of $\mathcal{M}$ prompts.

\paragraph{Encoder}

We first use BERT \citep{devlin-etal-2019-bert} to encode the input sequence $\mathcal{T}$: 
\begin{equation*}
    \mathbf{H}^\mathcal{T} = \operatorname{BERT}\left(\mathcal{T}\right)
\end{equation*}

\noindent Note that in order to encode the sentence $X$ independent of the prompts, we block the attention of the prompts to the sentence by a prompt-agnostic attention mask, which has a lower left submatrix of size $n\times k$ as a full $-inf$ matrix, where $k$ is the length of the prompt sequence. Then by indexing on the corresponding position of $\mathbf{H}^\mathcal{T}$, we can obtain the encoding of the sentence $X$ and the encodings of the two types of slots, denoted as $\mathbf{H}^X$, $\mathbf{H}^P$ and $\mathbf{H}^T$, where $\mathbf{H}^P, \mathbf{H}^T\in \mathbb{R}^{\mathcal{M}\times h}$ and $\mathbf{H}^X\in \mathbb{R}^{n\times h}$ and $h$ is the hidden size.

To enhance the interaction of different prompts, we designed extra prompt interaction layers. Each interaction layer contains self-attention between slots with the same type (the key, query and value are the encodings of slots) and cross-attention from sentence to prompt slots (the query is the encodings of slots while the key and value are the sentence encodings). Thus the final encodings of position and type slots ($\delta\in\{P,T\}$) are computed as follows:
\begin{align*}
    \hat{\mathbf{H}}^\delta = \operatorname{PromptInteraction}\left(\mathbf{H}^\delta+\mathbf{E}_{id}, \mathbf{H}^X\right) 
\end{align*}

\noindent where $\mathbf{E}_{id} \in \mathbb{R}^{\mathcal{M}\times h} $ denote the learnable identity embeddings of $\mathcal{M}$ prompts, which can bind position slot and type slot within the same prompt.

\paragraph{Entity Decoding}
Now we can decode the corresponding entity for each prompt by prompt locating and prompt typing, i.e., filling the position slot and type slot of the prompt. For the $i$-th prompt, we put its type slot $\hat{\mathbf{H}}^T_i$ through a classifier and get the probabilities of different entity types as:

\begin{equation*}
    \mathbf{p}^{t}_i = \operatorname{Classifier}\left( \hat{\mathbf{H}}^T_i\right)
\end{equation*}

\noindent where the classifier is a linear layer followed by the softmax function. For prompt locating, we need to determine whether the $j$-th word is the start or end word of the predicted entity by the $i$-th prompt. We first feed the position slot $\hat{\mathbf{H}}^P$ into a linear layer, and then add it with the word representation $\mathbf{H}^X$ of each position to get the fusion representations $\mathbf{H}^{F}$. We then perform binary classification to obtain the probability of the $j$-th word being the left boundary of the predicted entity for the $i$-th prompt:
\begin{align*}
    \mathbf{H}^{F} = \mathbf{W}_1\hat{\mathbf{H}}^P_i+\mathbf{W}_2\mathbf{H}^X \\
    \mathbf{p}^{l}_{ij} = \operatorname{Sigmoid}\left(\mathbf{W}_3\mathbf{H}^{F}_{ij}\right)
\end{align*}

\noindent where $\mathbf{W}_1, \mathbf{W}_2, \mathbf{W}_3\in \mathbb{R}^{h\times h}$ are learnable weights. In the same way, we can compute the probability $\mathbf{p}^{r}_{ij}$ of the $j$-th word being the right boundary. Then the probabilities of the entities predicted by the $\mathcal{M}$ prompts can be denoted as $\hat{\mathbf{Y}} = \{\hat{\mathbf{Y}}_{i}\}^\mathcal{M}_{i=1}$, where $\hat{\mathbf{Y}}_{i} = (\mathbf{p}^{l}_{i}, \mathbf{p}^{r}_{i}, \mathbf{p}^{t}_i)$ \footnote{\;$\mathbf{p}^{\alpha }_{i} = \left[\mathbf{p}^{\alpha }_{i0}, \mathbf{p}^{\alpha }_{i1}, \dots, \mathbf{p}^{\alpha }_{iN} \right] $, where $\alpha \in \{l, r\}$}.

\paragraph{Inference} During inference, we can get the left boundary, right boundary and type of the entity corresponding to the $i$-th prompt as $\left(\operatorname{argmax}\mathbf{p}^{l}_{i},\operatorname{argmax}\mathbf{p}^{r}_{i},\operatorname{argmax}\mathbf{p}^{t}_{i}\right)$. When two prompts yield identical entities, we keep only one; for conflicting candidates, such as entities with the same location but inconsistent types, we keep the entity with the highest probability.

\subsection{Dynamic Template Filling}
\label{3.4}

Since the correspondence between prompts and entities is unknown, we cannot assign labels to the slots in advance.
To solve it, we treat slot filling as a linear assignment problem\footnote{\;\url{https://en.wikipedia.org/wiki/Assignment_problem}}, where any entity can be filled to any prompt, incurring a cost, and we need to get the correspondence between the prompts and the entities with minimum overall cost. We propose a dynamic template filling mechanism to perform bipartite graph matching between prompts and the entities.
Let us denote the gold entities as $\mathbf{Y} = \{(l_i,r_i,t_i)\}^K_{i=1}$, where $K$ denotes the number of entities and $l_i,r_i,t_i$ are the boundary indices and type for the $i$-th entity. We pad $\mathbf{Y}$ with $\varnothing$ to ensure that it has the same number $\mathcal{M}$ as the prompts. Then the permutation of the prompts corresponding to the optimal match is:
\begin{equation*}
\label{cost}
\sigma^{\star}=\underset{\sigma \in \mathfrak{S}(\mathcal{M})}{\arg \min } \sum_{i=1}^{\mathcal{M}} \mathcal{C}ost_{m a t c h}\left(\mathbf{Y}_{i}, \hat{\mathbf{Y}}_{\sigma(i)}\right)
\end{equation*}

\noindent where $\mathfrak{S}(\mathcal{M})$ is the set of all $\mathcal{M}$-length permutations and  $\mathcal{C}ost_{m a t c h}\left(\mathbf{Y}_{i}, \hat{\mathbf{Y}}_{\sigma(i)}\right)$ is the pairwise match cost between the $i$-th entity and the prediction by the $\sigma(i)$-th prompt, we define it as $-\mathds{1}_{\left\{t_{i} \neq \varnothing\right\}} \left[ \mathbf{p}^{t}_{\sigma(i)}\left(t_{i}\right) + \mathbf{p}^{l}_{\sigma(i)}\left(l_{i}\right) + \mathbf{p}^{r}_{\sigma(i)}\left(r_{i}\right) \right]$, where $\mathds{1}_{\{\cdot\}}$ denotes an indicator function. 

Traditional bipartite graph matching is one-to-one, with each gold entity matching only one prompt, which leads to many prompts being matched to $\varnothing$, thus reducing the training efficiency. To improve the utilization of prompts, we extend the one-to-one bipartite graph matching to one-to-many, which ensures that a single gold entity can be matched by multiple prompts. To perform one-to-many matching, we simply repeat the gold entities to augment $\mathbf{Y}$ under a predefined upper limit $U$. In our experiments, we take $U=0.9\mathcal{M}$.
We use the Hungarian algorithm \citep{kuhn1955hungarian} to solve Equation \ref{cost} for the optimal matching $\sigma^\star$ at minimum cost. Then the losses for prompt locating ($\mathcal{L}_2$) and typing ($\mathcal{L}_1$) are computed as follows:
\begin{equation*}
\mathcal{L}_{1} =-\sum_{i=1}^{\mathcal{M}}\log \mathbf{p}_{\sigma^{\star}(i)}^{t}\left(t_{i}\right)
\end{equation*}
\begin{equation*}
\mathcal{L}_{2} =-\sum_{i=1}^{\mathcal{M}}\underset{t_{i} \neq \varnothing}{\mathds{1}}{}\left[\log \mathbf{p}_{\sigma^{\star}(i)}^{l}\left(l_{i}\right)+\log \mathbf{p}_{\sigma^{\star}(i)}^{r}\left(r_{i}\right)\right]
\end{equation*}

\noindent and the final loss is the weighted sum $\mathcal{L} = \lambda_1\mathcal{L}_1+\lambda_2\mathcal{L}_2$. By default we set $\lambda_1 = 1 $ and $\lambda_2 =2$.

\begin{table*}[]
\centering
\small
\begin{tabular}{lccccccccccc}
\toprule
\multirow{2}{*}{Model}   & \multicolumn{3}{c}{ACE04} & \multicolumn{3}{c}{ACE05}   & \multicolumn{3}{c}{CoNLL03} \\

 \cmidrule(lr){2-4}  \cmidrule(lr){5-7} \cmidrule(lr){8-10} 
&  Pr.  & Rec. & F1 & Pr.  & Rec. & F1  & Pr.  & Rec. & F1  \\
\midrule
Biaffine \citep{yu-etal-2020-named}       & 87.30  & 86.00  & 86.70 & 85.20  & 85.60  & 85.40  & 93.70 & 93.30 & 93.50 \\
MRC \citep{li-etal-2020-unified} & 85.05 & 86.32 &  85.98 & 87.16 & 86.59 & 86.88 & 92.33 & 94.61 & 93.04\\
BARTNER \citep{yan-etal-2021-unified-generative} & 87.27 & 86.41 & 86.84 & 83.16 & 86.38 & 84.74 & 92.61  & 93.87 & 93.24  \\
Seq2Set \citep{ijcai2021-542} & 88.46 & 86.10 & 87.26 & 87.48 & 86.63 & 87.05 & - & - & - \\
Triaffine \citep{yuan-etal-2022-fusing} & 87.13 &  87.68 &  87.40 &  86.70 &  86.94 &  86.82 & - & - & - \\
UIE \citep{lu-etal-2022-unified} & - & - & 86.89 & - & - & 85.78 & - & - & 92.99 \\
W$^{2}$NER \citep{li2022unified} & 87.33 &  87.71 &  87.52 & 85.03 & 
88.62 &  86.79 & 92.71 &  93.44 &  93.07 \\
BuParser\citep{yang-tu-2022-bottom} & 86.60  &  87.28   & 86.94  &  84.61  &  86.43  &  85.53  & - & - & - \\
LLCP \cite{lou-etal-2022-nested} & 87.39 &  88.40 &  87.90 &  85.97 &  87.87 &  86.91  & - & - & - \\
PIQN \citep{shen-etal-2022-parallel} & 88.48 & 87.81 & 88.14 & 86.27 & 88.60 & 87.42 & 93.29 & 92.46 & 92.87  \\
BS [BERT-large] \citep{zhu-li-2022-boundary} & - & - &  87.85 & - &  - &  87.82 &  - &  - &  {93.08} \\
BS [RoBERTa-large] \citep{zhu-li-2022-boundary} & - & - &  88.52 & - &  - &  88.14 &  - &  - &  \textbf{93.77} \\
\midrule
PromptNER [BERT-large] & 87.58 & 88.76 & 88.16 & 86.07 & 88.38 & 87.21  & 92.48 & 92.33 & 92.41 \\
PromptNER [RoBERTa-large] & \textbf{88.64} & \textbf{88.79} & \textbf{88.72} & \textbf{88.15} & {88.38} & \textbf{88.26} & 92.96 & 93.18 & 93.08\\
\bottomrule
\end{tabular}
\caption{Results in the standard \textit{flat} and \textit{nested} NER setting.}
\label{tab:sup}
\end{table*}

\section{Experiments}

To verify the effectiveness of PromptNER in various settings, we conduct extensive experiments in flat and nested NER (\Cref{flatnested}) and low-resource NER, including in-domain few-shot setting (\Cref{indomain}) and cross-domain few-shot setting (\Cref{crossdomain}).

\subsection{Implementation Details} If not marked, we use BERT-large \citep{devlin-etal-2019-bert} as the backbone of the model. We use reserved tokens and sparse tokens of BERT, e.g. \texttt{[unused1]}-\texttt{[unused100]}, as position and type slots. 
The model has a hidden size $h = 1024$ and $\mathcal{I}=3$ prompt interaction layers. 
Since the maximum number of entities per sentence does not exceed 50, we uniformly set the number of prompts $\mathcal{M} = 50$. In the dynamic template filling mechanism, we set the upper limit of the expanded labels $U=0.9\mathcal{M}=45$ for extended bipartite graph matching. For all datasets, we train PromptNER for 50-100 epochs and use the Adam \citep{adam}, with a linear warmup and linear decay learning rate schedule and a peak learning rate of 2$e$-5. We initialize our prompt identity embeddings $\mathbf{E}_{id}$ with the normal distribution $\mathcal{N}(0.0,0.02)$.

\subsection{Warmup Training}

Before employing PromptNER in the low-resource scenario, we use the open Wikipedia data to warm up the training for entity locating. PromptNER needs to elicit the language model to locate entities, while the pre-trained language model does not learn entity localization during pre-training. Therefore PromptNER needs to learn the prompt locating ability initially by Wiki warmup training. We choose accessible Wikipedia as our warm-up training data. Wikipedia contains a wealth of entity knowledge \citep{yamada-etal-2020-luke, wang2022damonlp} that is useful for entity-related tasks such as named entity recognition, relation extraction, entity linking, etc. 
We call entity-related hyperlinks in Wikipedia as wiki anchors.
These anchors only have position annotations and lack type information, and we use these partially annotated noisy data to warm up the localization ability of PromptNER. Specifically, we fix the weight of BERT, train 3 epochs with a learning rate of 1$e$-5 on the constructed wiki anchor data, and optimize the model only on the entity locating loss to warm up the entity decoding module. In low-resource scenarios (in-domain few-shot setting in \cref{indomain} and cross-domain few-shot setting in \cref{crossdomain}), we initialize PromptNER with the warmed-up weights.

\subsection{Standard Flat and Nested NER Setting}
\label{flatnested}

\paragraph{Datasets} We adopt three widely used datasets to evaluate the performance of the model in the standard NER setting, including one flat NER dataset: CoNLL03 \citep{tjong-kim-sang-de-meulder-2003-introduction} and two nested NER datasets: ACE04 \citep{ doddington-etal-2004-automatic} and ACE05 \citep{2005-automatic}. For ACE04 and ACE05, we use the splits of \citet{lu-roth-2015-joint,muis-lu-2017-labeling} and the preprocessing protocol of \citet{shibuya-hovy-2020-nested}. Please refer to Appendix \ref{app:statistics} for detailed statistics on nested entities about ACE04 and ACE05. For CoNLL03, we follow \citet{lample-etal-2016-neural,yu-etal-2020-named, 10.1007/978-3-031-30675-4_31} to train the model on the concatenation of the train and dev sets.

\paragraph{Baselines} We select recent competitive models as our baseline, including span-based \citep{yuan-etal-2022-fusing, li2022unified}, generation-based \citep{ijcai2021-542, yan-etal-2021-unified-generative, lu-etal-2022-unified}, MRC-based \citep{li-etal-2020-unified,shen-etal-2022-parallel, jwq2022rcekgqa}, and parsing-based \citep{yu-etal-2020-named, zhu-li-2022-boundary, lou-etal-2022-nested, yang-tu-2022-bottom}. These methods adopt different pre-trained language models as the encoder, thus in the experimental results, we provide the performance of PromptNER on BERT-large and RoBERTa-large.

\paragraph{Results}

Table \ref{tab:sup} illustrates the performance of PromptNER as well as baselines on the flat and nested NER datasets. We observe that PromptNER outperforms most of the recent competitive baselines. When using RoBERTa-large as the encoder, PromptNER outperforms previous state-of-the-art models on the nested NER datasets, achieving F1-scores of 88.72\% and 88.26\% on ACE04 and ACE05 with +0.20\% and +0.12\% improvements. And on the flat NER dataset CoNLL03, PromptNER achieves comparable performance compared to the strong baselines. We also evaluate the performance of entity locating and entity typing separately on ACE04, please refer to Appendix \ref{app:analysissubtask}.

\subsection{In-Domain Few-Shot NER Setting}
\label{indomain}

\begin{table}[]
    \centering
    \small
    \begin{tabular}{>{\centering\arraybackslash}p{1.8cm}>{\centering\arraybackslash}p{0.60cm}>{\centering\arraybackslash}p{0.60cm}>{\centering\arraybackslash}p{0.60cm}>{\centering\arraybackslash}p{0.60cm}c}
    \toprule
         Models & ORG & PER & LOC$^\star$ & MISC$^\star$ & Overall \\
         \midrule
         BERTTagger & 75.32 & 76.25  & 61.55 & 59.35 & 68.12 \\
         TemplateNER  & 72.61 & 84.49 & 71.98 &  \textbf{73.37} & 75.59 \\
         \midrule
         PromptNER & \textbf{76.96}  & \textbf{88.11} & \textbf{82.69}  & 62.89  & \textbf{79.75} \\
         \bottomrule
    \end{tabular}
    \caption{Results in the in-domain few-shot NER setting. \quad \quad $^\star$ indicates the low-resource entity type.}
    \label{tab:indomainfewshot}
\end{table}

\begin{table*}[!h]
\centering
\small
\begin{tabular}{l>{\centering\arraybackslash}p{0.4cm}>{\centering\arraybackslash}p{0.4cm}>{\centering\arraybackslash}p{0.4cm}>{\centering\arraybackslash}p{0.4cm}>{\centering\arraybackslash}p{0.4cm}>{\centering\arraybackslash}p{0.4cm}>{\centering\arraybackslash}p{0.4cm}>{\centering\arraybackslash}p{0.4cm}>{\centering\arraybackslash}p{0.4cm}>{\centering\arraybackslash}p{0.4cm}>{\centering\arraybackslash}p{0.4cm}>{\centering\arraybackslash}p{0.4cm}>{\centering\arraybackslash}p{0.4cm}>{\centering\arraybackslash}p{0.4cm}>{\centering\arraybackslash}p{0.4cm}>{\centering\arraybackslash}p{0.4cm}}
\toprule
 \multirow{2}{*}{Methods}  & \multicolumn{6}{c}{{MIT Movie}} & \multicolumn{6}{c}{{MIT Restaurant}} & \multicolumn{3}{c}{{ATIS}} & \multirow{2}{*}{Avg.} \\
 \cmidrule(lr){2-7} \cmidrule(lr){8-13} \cmidrule(lr){14-16}
& 10 & 20 &  50 &  100 & 200 & 500 & 10 & 20 &  50 &  100 & 200 & 500  & 10 & 20 &  50  \\
\midrule
NeighborTagger& 3.1 & 4.5 & 4.1 & 5.3 & 5.4 & 8.6 & 4.1 &  3.6  & 4.0  & 4.6 & 5.5 &  8.1 & 2.4 &  3.4 &  5.1 & 4.8\\
Example-based & 40.1 & 39.5 & 40.2 & 40.0 & 40.0 & 39.5 & 25.2  & 26.1 &  26.8  & 26.2 &  25.7 &  25.1 & 22.9 &  16.5 &  22.2 & 30.4\\
MP-NSP  & 36.4 & 36.8 & 38.0 & 38.2 & 35.4 & 38.3 & 46.1 &  48.2 &  49.6 &  49.6  & 50.0  & 50.1 & 71.2 &  74.8 &  76.0 & 49.2\\
BERTTagger & 28.3 &  45.2 &  50.0 &  52.4 &  60.7 &  76.8 & 27.2 & 40.9 & 56.3 & 57.4 & 58.6 &  75.3 & 53.9 & 78.5 & 92.2 & 56.9\\
TemplateNER  & 42.4 & 54.2 & 59.6 & 65.3 & 69.6 & 80.3 & 53.1 &  60.3 &  64.1 &  67.3 &  72.2 &  75.7 & 77.3 &  88.9 &  93.5 & 68.3 \\
\midrule
PromptNER & \textbf{55.6} & \textbf{68.2} & \textbf{76.5} & \textbf{80.4} & \textbf{82.9}  & \textbf{84.5} & \textbf{56.1} & \textbf{62.6} & \textbf{69.3} & \textbf{71.3} & \textbf{74.4} & \textbf{77.4} & \textbf{91.5} & \textbf{94.3} & \textbf{95.5} & \textbf{76.0} \\
\bottomrule
\end{tabular}

\caption{Results in the cross-domain few-shot NER setting. We transfer the model from the general domain (CoNLL03) to specific target domains with only a few labeled instances: Movie Review, Restaurant Review, ATIS.}
\label{tab:crossdomainfewshot}
\end{table*}

\paragraph{Datasets and Baselines} Following \citet{cui-etal-2021-template}, we construct a dataset with low-resource scenarios based on CoNLL03. We limit the number of entities of specific types by downsampling and meet the low-resource requirement on these types. Specifically, we set \texttt{LOC} and \texttt{MISC} as low-resource types and \texttt{PER} and \texttt{ORG} as resource-rich types. We downsample the CoNLL03 training set to obtain 4,001 training samples, including 100 \texttt{MISC}, 100 \texttt{LOC}, 2496 \texttt{PER}, and 3763 \texttt{ORG} entities. 
We use this dataset to evaluate the performance of PromptNER under the in-domain few-shot NER setting. We choose BERTTagger \citep{devlin-etal-2019-bert} and the low-resource friendly model TemplateNER \citep{cui-etal-2021-template} as our baselines.

\paragraph{Results} As shown in Table \ref{tab:indomainfewshot}, we achieve significant performance improvements on both low and rich resource types compared to BERTTagger. In particular, we achieve an average +12.34\% improvement on low-resource types. Prompt design is the key to prompt learning \citep{https://doi.org/10.48550/arxiv.2107.13586}, and our method adaptively learns them by the dynamic template filling mechanism which can achieve better performance in low resource scenarios. Compared to TemplateNER, PromptNER performs better in the low-resource \texttt{LOC} type and overall, and slightly weaker in \texttt{MISC} type.
We believe that entities of type \texttt{MISC} are more diverse and it is hard for PromptNER to learn a clear decision boundary from a small number of support instances.

\subsection{Cross-Domain Few-Shot NER Setting}
\label{crossdomain}

\paragraph{Datasets and Baselines}
In practical scenarios, we can transfer the model from the resource-rich domain to enhance the performance of the low-resource domain.
In this setting, the entity types of the target domain are different from the source domain, and only a small amount of labeled data is available for training. To simulate the cross-domain few-shot setting, we set the source domain as the resource-rich CoNLL03 dataset, and randomly sample some training instances from the MIT movie, MIT restaurant, and ATIS datasets as the training data for the target domain. Specifically, we randomly sample a fixed number of instances for each entity type (10, 20, 50, 100, 200, 500 instances per entity type for MIT movie and MIT restaurant, and 10, 20, 50 instances per entity type for ATIS). If the number of instances of a type is less than the fixed number, we use all instances for training. We select several competitive methods with the same experimental setup as our baselines, including NeighborTagger \citep{wiseman-stratos-2019-label}, Example-based \citep{ DBLP:journals/corr/abs-2008-10570}, MP-NSP \citep{huang-etal-2021-shot}, BERTTagger \citep{devlin-etal-2019-bert}, and TemplateNER \citep{ cui-etal-2021-template}.

\paragraph{Results}

Table \ref{tab:crossdomainfewshot} shows the performance of PromptNER in the cross-domain few-shot setting, along with some strong baselines. We observe that PromptNER achieves the best performance on all settings of fixed support instances for the three datasets. At the extreme 10-shot setting, PromptNER outperforms TemplateNER by +13.2\%, +3\%, and +14.2\% on the MIT Movie, MIT Restaurant, and ATIS datasets, respectively. Overall, compared to the previous state-of-the-art model, PromptNER achieves a +7.7\% improvement on average in all cross-domain few-shot settings. 
This shows that PromptNER can transfer the generalized knowledge learned in the resource-rich domain to the low-resource domain. Furthermore, PromptNER can decouple entity locating and typing via position and type slots, which is especially suitable for cross-domain scenarios with syntactic consistency and semantic inconsistency.

\section{Analysis}

\subsection{Ablation Study}

\begin{table}[]
    \centering
    \small
    \begin{tabular}{lccc}
    \toprule
    Model & Pr. & Rec. & F1\\
    \midrule
    \textsc{Default} &   \textbf{87.58} &  \textbf{88.76} &  \textbf{88.16} \\
    \midrule
    \quad \textit{w/o Dyn. Template Filling}  & 86.19 & 83.32 & 84.73 \\
    \quad \textit{w/o Extended Labels} & 84.46  &  83.65 & 84.05  \\
    \quad \textit{w/o Prompt-agnostic Mask} & 87.59 & 87.90 & 87.74\\
    \bottomrule
    \end{tabular}
    \caption{Ablation study. (1) \textit{w/o Dyn. Template Filling}: static template filling, filling the slots of the prompts according to the occurrence order of the entities; (2) \textit{w/o Extended Labels}: no label expansion in the dynamic template filling mechanism, i.e., using the traditional one-to-one bipartite graph matching; (3) \textit{w/o Prompt-agnostic Mask}: using the original BERT for encoding.}
    \label{tab:ablation}
\end{table}

We conduct ablation experiments on ACE04 to analyze the effect of different modules of PromptNER. The experimental results are shown in Table \ref{tab:ablation}, without the three practices, there is a different degradation of model performance.
If we assign labels to slots simply by entity order or use one-to-one bipartite graph matching, the model performance decreases by 3.43\% and 4.11\%, respectively.
We conclude that a one-to-many dynamic template-filling mechanism is important as it allows prompts fit to related entities adaptively. The one-to-many manner ensures that an entity can be predicted by multiple prompts, improving the model prediction tolerance. 
When encoding the input sequence, it is also important to keep the sentence encoding to be prompt agnostic, resulting in a +0.42\% performance improvement.

\subsection{Analysis of $\mathcal{M}$ and $\mathcal{I}$}

\begin{figure}[!h]
    \centering
    \includegraphics[width=1\linewidth]{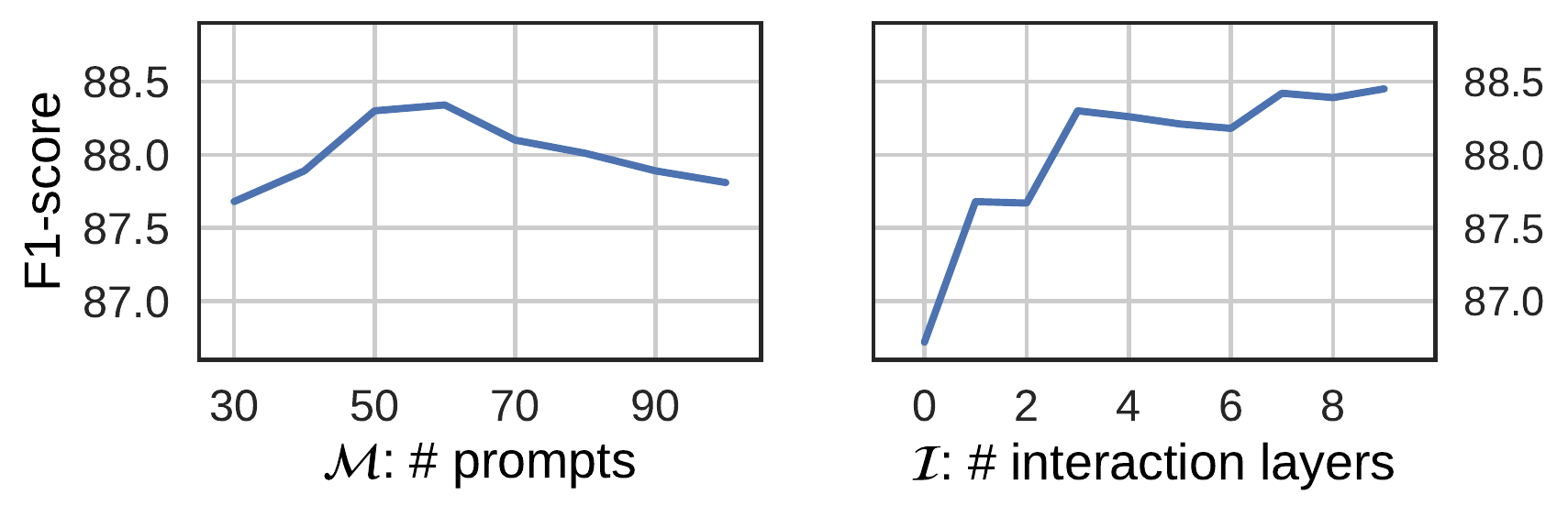}
    \caption{F1-scores under different number of prompts $\mathcal{M}$ and interaction layers $\mathcal{I}$ on ACE04 dataset}
    \label{fig:mi}
\end{figure}

We further investigate the effect of the number of prompts and the number of prompt interaction layers on PromptNER.
From Figure \ref{fig:mi}, we can observe that the number of prompts is more appropriate between 50 and 60. Too few would make it difficult to cover all entities, and too many would exceed the maximum length of the encoding and impair the model performance.
In addition, as the number of interaction layers increases, we can observe a significant performance improvement in Figure \ref{fig:mi}.
This suggests that the interaction between prompts can model the connection between entities.
Considering the size and efficiency of the model, we choose $\mathcal{M}$=50, $\mathcal{I}$=3 as the default setting.

\subsection{Analysis of Prompt Templates}
\label{template}
Templates are important for prompt learning \citep{gao-etal-2021-making, ding-etal-2021-nerd}. In this section, we conduct experiments on ACE04 to analyze the effect of different templates, as shown in Table \ref{tab:template}. Contrary to intuition, inserting hard or soft contextual tokens in the prompts does not improve the model performance. We argue that adding contextual tokens to our multi-prompt template significantly grows the input sequence (each additional token increases the total length by $\mathcal{M}$), and the long sequence may exceed the maximum encoding length of BERT. Comparing hard and soft templates, we find that soft templates are more useful, which is consistent with \citet{ding-etal-2021-nerd}.

\begin{table}[!htp]
    \centering
    \small
    \begin{tabular}{>{\arraybackslash}p{0.7cm}>{\arraybackslash}p{4.1cm}c}
    \toprule
        Type & Template &  F1 \\ 
    \midrule
Hard & \multicolumn{1}{m{4.9cm}} {\{\texttt{[P$_i$]}\textit{ is a }\texttt{[T$_i$]}\}$_{i=1,2,\cdots,50}$\textit{ entity }\texttt{[CLS]}\textit{Jobs was born in San Francisco.}}  & 87.96 \\
\midrule
Soft & \multicolumn{1}{m{4.9cm}} {\{\texttt{[P$_i$]}\texttt{<s>}\texttt{[T$_i$]}\}$_{i=1,2,\cdots,50}$\texttt{[CLS]}\textit{Jobs was born in San Francisco.}}  & 88.05 \\
\midrule
Default & \multicolumn{1}{m{4.9cm}} {\{\texttt{[P$_i$]} \texttt{[T$_i$]}\}$_{i=1,2,\cdots,50}$ \texttt{[CLS]}\textit{ Jobs was born in San Francisco.}}  &  88.16 \\
    \bottomrule
    \end{tabular}
    \caption{A comparison of different templates. \texttt{<s>} is a learnable contextual token and the default template contains only slots without any contextual tokens.}
    \label{tab:template}
\end{table}

\subsection{Inference Efficiency}
Theoretically, for a sentence with $N$ words and $C$ potential entity types, type-oriented \citep{li-etal-2020-unified} and span-oriented \citep{cui-etal-2021-template} prompt learning need to be run $C$ and $N(N-1)/2$ times.
And the generation-based methods \citep{yan-etal-2021-unified-generative} generate entity sequences in an autoregressive manner. Assuming that the length of the entity sequence is $T$, it takes $T$ steps to decode all entities.
In contrast, PromptNER can locate and typing the entities in parallel through dual-slot multi-prompt learning, it only needs one run to decode all the entities. Under the same experimental setup, we compare their inference efficiency on CoNLL03, as shown in Table \ref{tab:inference}. Empirically, PromptNER achieves the fastest inference efficiency compared to the baselines, with 48.23$\times$, 1.86$\times$ and 2.39$\times$ faster than TemplateNER, MRC and BARTNER, respectively.

\begin{table}[!htp]
    \centering
    \small
    \begin{tabular}{lcc}
    \toprule
    Model & Complexity & SpeedUp \\
    \midrule
     {TempNER \citep{cui-etal-2021-template}} & $O(N^2)$ & 1.00$\times$ \\ 
     {MRC \citep{li-etal-2020-unified}}  & $O(C)$ & 25.86$\times$  \\  
     {BARTNER \citep{yan-etal-2021-unified-generative}} & $O(T)$  &  20.17$\times$ \\  
     
     \midrule
     PromptNER & $O(1)$ & 48.23$\times$  \\  
    \bottomrule
    \end{tabular}
    \caption{A comparison of inference efficiency on the test set of CoNLL03. All experiments were conducted with one NVIDIA GeForce RTX 3090 graphics card.}
    \label{tab:inference}
\end{table}

\section{Conclusion}

In this paper, we unify entity locating and entity typing in prompt learning for NER with a dual-slot multi-prompt template. By filling position slots and type slots, our proposed model can predict all entities in one round. We also propose a dynamic template filling mechanism for label assignment, where the extended bipartite graph matching assigns labels to the slots in a one-to-many manner. We conduct extensive experiments in various settings including flat and nested NER and low-resource in-domain and cross-domain NER, and our model achieves superior performance compared to the competitive baselines. 

\section*{Limitations}

We discuss here the limitations of the proposed PromptNER. First, although PromptNER performs well on flat and nested NER, it cannot recognize discontinuous entities. The discontinuous entity can be divided into multiple fragments, while each position slot of PromptNER can only fill one. A simple alternative is to expand the position slots in prompts to accommodate discontinuous entities. Second, named entity recognition requires pretrained language models (PLMs) with the essential ability to sense the structure and semantics of entities, which can enhance entity locating and entity typing in low-resource scenarios. However, since PLMs prefer to learn semantic rather than structured information in the pre-training stage, PromptNER needs to be warmed up by Wiki training when applied to low-resource scenarios. Finally, since the number of prompts is determined during training, there is a limit to the number of entities that the model can recognize. If the number of entities in a sentence exceeds the pre-specified value when testing, PromptNER will perform poorly.

\section*{Acknowledgments}
This work is supported by the Key Research and Development Program of Zhejiang Province, China (No. 2023C01152, No. 2022C01011), the Fundamental Research Funds for the Central Universities (No. 226-2023-00060), and MOE Engineering Research Center of Digital Library.

\bibliography{anthology,custom}
\bibliographystyle{acl_natbib}

\clearpage

\appendix

\section{Appendix}

\subsection{Statistics of the nested NER datasets}
\label{app:statistics}

In Table \ref{tab:statistics}, we present statistics for the standard nested datasets: ACE04 and ACE05. We report the number of sentences (\#S), the number of sentences containing nested entities (\#NS), the average sentence length (AL), the number of entities (\#E), the number of nested entities (\#NE), the nesting rate (NR), and the maximum and the average number of entities (\#AE) in sentences on the two datasets.

\begin{table}[h!]
    \centering
    \small
\begin{tabular}{>{\centering\arraybackslash}p{0.6cm}>{\centering\arraybackslash}p{0.6cm}>{\centering\arraybackslash}>{\centering\arraybackslash}p{0.6cm}>{\centering\arraybackslash}p{0.65cm}>{\centering\arraybackslash}p{0.65cm}>{\centering\arraybackslash}p{0.6cm}>{\centering\arraybackslash}p{0.6cm}}
\toprule
& \multicolumn{3}{c}{ ACE04 } & \multicolumn{3}{c}{ ACE05 } \\
\cmidrule(lr){ 2 -4 } \cmidrule(lr){5-7} & Train & Dev & Test & Train & Dev & Test \\
\midrule \#S & 6198 & 742 & 809 & 7285 & 968 & 1058 \\
\#NS & 2718 & 294 & 388 & 2797 & 352 & 339 \\
\#E & 22204 & 2514 & 3035 & 24827 & 3234 & 3041 \\
\#NE & 10159 & 1092 & 1417 & 10039 & 1200 & 1186 \\
NR & 45.75 & 43.44 & 46.69 & 40.44 & 37.11 & 39.00 \\
AL & 21.41 & 22.13 & 22.03 & 18.82 & 18.77 & 16.93 \\
\#ME & 28 & 22 & 20 & 28 & 23 & 20 \\
\#AE & 3.58 & 3.38 & 3.75 & 3.41 & 3.34 & 2.87 \\
\bottomrule
\end{tabular}
    \caption{Statistics for ACE04 and ACE05 datasets. }
    \label{tab:statistics}
\end{table}

\subsection{Analysis of Entity Locating and Typing}
\label{app:analysissubtask}

Our work unifies entity locating and entity typing in prompt learning, and in this section we compare the performance of the model on the two subtasks with some strong baselines. Following \citet{shen-etal-2022-parallel}, we consider entity locating correct when the left and right boundaries are correctly predicted. Based on the accurately located entities, we then evaluate the performance of entity typing. Figure \ref{tab:loctyp} shows the performance comparison on ACE04, PromptNER significantly outperforms the baseline for both tasks, achieving +0.59\% and +0.56\% improvement in entity locating and entity typing compared to \citet{shen-etal-2022-parallel}.

\begin{table}[h!]
\centering
\small
\begin{tabular}{l>{\centering\arraybackslash}p{0.6cm}>{\centering\arraybackslash}p{0.6cm}>{\centering\arraybackslash}p{0.6cm}}

\toprule
Model & Pr.  & Rec. & F1  \\
\midrule
\multicolumn{4}{c}{{Entity Locating}} \\
\midrule
Seq2set \citep{ijcai2021-542} &  {92.75}  &      90.24    &    91.48    \\
Locate\&label \citep{shen-etal-2021-locate} & 92.28  &      90.97   &     91.62  \\
PIQN \citep{shen-etal-2022-parallel}    & {92.56}      &  {91.89}    &   {92.23}  \\
\midrule
PromptNER & { 91.86 } & \textbf{93.80} &   \textbf{92.82} \\
\midrule
\multicolumn{4}{c}{{Entity Typing}} \\
\midrule
 Seq2set \citep{ijcai2021-542}  & 95.36   &     86.03      &  90.46    \\
Locate\&label \citep{shen-etal-2021-locate} & 95.40  &      86.75   &     90.87  \\
PIQN \citep{shen-etal-2022-parallel}   &  {95.59}    &     {87.81}     &    {91.53} \\
\midrule
PromptNER & 95.15  &  \textbf{89.22}  &  \textbf{92.09}  \\
\bottomrule
\end{tabular}
\caption{Analysis of entity locating and typing.}
\label{tab:loctyp}
\end{table}

\end{document}